\documentclass[lettersize,journal]{IEEEtran}
\ifCLASSINFOpdf
\else
\fi

\usepackage{graphicx}
\usepackage{multirow}
\usepackage{color}
\usepackage{diagbox}
\usepackage{amsmath}
\usepackage{amssymb}

\begin{document}

\title{A Unified Framework to Super-Resolve Face Images of Varied Low Resolutions}

\author{Qiuyu~Peng,
        Zifei~Jiang,
        Yan~Huang
        and~Jingliang~Peng%,~\IEEEmembership{Member,~IEEE}% <-this % stops a space
\IEEEcompsocitemizethanks{

\IEEEcompsocthanksitem Q. Peng, Z. Jiang and Y. Huang are all with School of Software, Shandong University, Jinan, Shandong, 250101, P. R. China.
J. Peng is with Shandong Provincial Key Laboratory of Network Based Intelligent Computing \& School of Information Science and Engineering, University of Jinan, Jinan, Shandong, 250022, P. R. China. 
%%JLP_TMM: added
Y. Huang and J. Peng are the corresponding authors.\protect\\
E-mails: pngqiuyu@mail.sdu.edu.cn, jiangzifei@mail.sdu.edu.cn, yan.h@sdu.edu.cn, ise\_pengjl@ujn.edu.cn.
}% <-this % stops an unwanted space
\thanks{}}

%{Shell %\MakeLowercase{\textit{et al.}}: A Sample Article Using IEEEtran.cls for IEEE Journals}

\maketitle
\def\eg{\emph{e.g.}}
\def\Eg{\emph{E.g.}}
\def\etal{\emph{et al.}}
\def\ie{\emph{i.e.}}
\def\etc{\emph{etc.}}

\def\peng{\textcolor[rgb]{0.8,0.1,0.1}}

\begin{abstract}
The existing face image super-resolution (FSR) algorithms usually train a specific model for a specific low input resolution for optimal results. By contrast, we explore in this work a unified framework that is trained once and then used to super-resolve input face images of varied low resolutions. For that purpose, we propose a novel neural network architecture that is composed of three anchor auto-encoders, one feature weight regressor and a final image decoder. The three anchor auto-encoders are meant for optimal FSR for three pre-defined low input resolutions, or named anchor resolutions, respectively. An input face image of an arbitrary low resolution is firstly up-scaled to the target resolution by bi-cubic interpolation and then fed to the three auto-encoders in parallel. 
%An input face image of an arbitrary low resolution is firstly fed to the three auto-encoders in parallel. 
The three encoded anchor features are then fused with weights determined by the feature weight regressor. At last, the fused feature is sent to the final image decoder to derive the super-resolution result. As shown by experiments, the proposed algorithm achieves robust and state-of-the-art performance over a wide range of low input resolutions by a single framework. Code and models will be made available after the publication of this work.
\end{abstract}

\section{Introduction}
\label{sec:intro}

While face images are the core type of data underlying many applications, \eg, video surveillance, face recognition, face alignment and face manipulation, the raw images were often acquired at low resolutions due to the limited picturing conditions. Therefore, it is often desired to raise the resolution of a raw face image to facilitate more successful applications. Correspondingly, the field of face image super-resolution (FSR), or sometimes by the name of face hallucination, has been intensively researched in recent years.

In general, it is a great challenge to faithfully restore the high resolution of a face image, particularly so as human eyes (and even computers) are sensitive to even minute distortions or artifacts in a reconstructed face image. In order to tackle this challenge, many algorithms have been proposed since the early 2000s. Initially, algorithms based on traditional machine learning techniques were proposed for the FSR task. Since 2015 or so, deep-learning-based FSR algorithms have become the main-stream, leading to great successes for typical up-scaling factors (\eg, $\times 8$ and $\times 4$).
  
However, one issue with the existing FSR algorithms is that they usually optimize a model separately for each specific up-scaling factor. For deep-learning-based methods, this means that a separate neural network (NN) model should be trained and maintained for each specific up-scaling factor. This brings high computing and memory costs for training and maintaining different models for different up-scaling factors. In real application scenarios, this problem is exacerbated as face images may be captured at arbitrarily low resolutions, making it impractical to train and maintain a model for every possible low resolution.

%In order to solve this issue, we explore super-resolving face images of arbitrary low resolutions in a unified way with a single model. Specifically, we propose in this work a novel neural network framework that is trained once and used to super-resolve input face images of varied low resolutions. The proposed framework is composed of three auto-encoder branches, one weight regressor and one image decoder. The auto-encoders anchor the input image to three features roughly corresponding to three pre-defined anchor resolutions, fuse the anchor features with weights determined by the weight regressor and produces the final result from the fused feature by the image decoder. Major contributions of this work reside in the following aspects.

In order to solve this issue, we explore super-resolving face images of arbitrary low resolutions in a unified way with a single model. Specifically, we propose in this work a novel neural network framework that is trained once and used to super-resolve input face images of varied low resolutions. The proposed method anchors the input image of an arbitrary resolution by auto-encoders to three features roughly corresponding to three pre-defined anchor resolutions, which are then fused and sent to an image decoder to derive the final result. Major contributions of this work reside in the following aspects.

\begin{itemize}

%\item \textbf{Identification of the model-resolution binding issue}. To the best of our knowledge, we explicitly point out the issue of strong binding of model and input image resolution and explore ways to solve this issue for the first time.
\item \textbf{Identification of the issue with rigid model-resolution binding}. To the best of our konwledge, for the problem of FSR, we explicitly point out the issue with rigid binding of optimal model and input image resolution and explore solutions for the first time.
%To the best of our knowledge, we explicitly point out the issue with rigid binding of optimal model and input image resolution and explore ways to solve this issue for the first time.

\item \textbf{Unified FSR framework for varied input resolutions}. We propose an original algorithm with a unified neural network framework that is trained for once and applied to super-resolve face images of varied resolutions.

\item \textbf{Robust and state-of-the-art (SOTA) FSR performance}.  As shown by experiments, the proposed algorithm achieves robust and state-of-the-art performance over a wide range of low input resolutions by a single generic model.
%; furthermore, the proposed algorithm achieves performance that is even comparable with specific benchmark models trained for specific input resolutions.
\end{itemize}

\begin{figure*}[htb]
\centering
\includegraphics[width=1.0\textwidth]{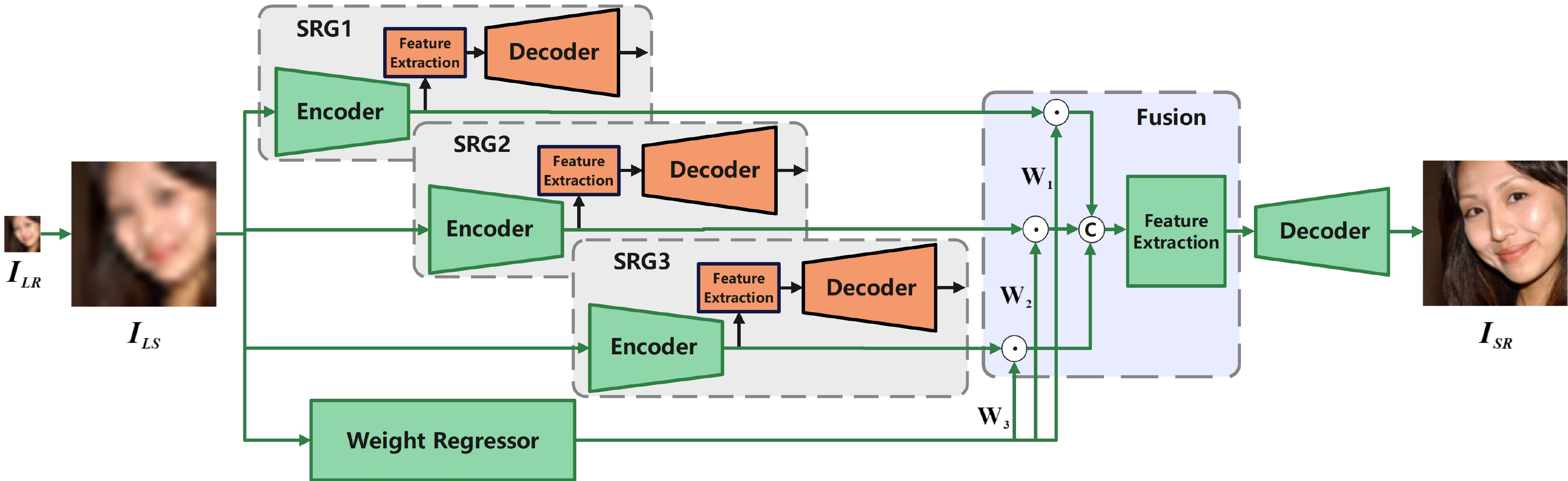} 
\caption{Complete architecture of the proposed UFVNet, including modules and data flows for both training and testing. It is composed of three image encoder-decoders (or named anchor SR generators, SRGs), a weight regressor, a fusion module and a final decoder module. All the blocks and arrows (indicating data flows) are used in the training,
%All the blocks and data flows (indicated by arrows) are used in the training, 
while only the green blocks and arrows are kept for the testing. }  

\label{fig:architecture}
\end{figure*}

\section{Related Work}

\subsection{Face Image Super-Resolution}

In early years, FSR algorithms~\cite{baker2000hallucinating,Liu2001ATA,Park2008AnEF,Yang2013StructuredFH,Farrugia2017FaceHU,Shi2019FaceHV,Zhuang2007HallucinatingFL,Huang2010SuperresolutionOH} are mostly based on traditional machine learning technology. In recent years, deep-learning-based approaches to FSR have achieved great successes. Zhou \etal~\cite{zhou2015learning} employ deep Bi-channel CNN for FSR. Yu and Porikli~\cite{yu2016ultra} use discriminative generative networks to generate realistic super-resolution (SR) images. Huang and Liu~\cite{huang2016face} develop a deep CNN for FSR with iterative back projection introduced for extra post-processing. Huang \etal~\cite{huang2017wavelet} propose a CNN to predict wavelet coefficients which are then used to reconstruct the high-resolution (HR) image. Jiang \etal~\cite{jiang2019atmfn} use multiple models to generate candidate SR results which are fused for the final result. Hu \etal~\cite{hu2019face} propose to predict a basic face and a compensation face and fuse them to derive the SR image. Lu \etal~\cite{lu2020global} propose a global-local fused network for FSR, paying special attention to high-frequency information. % for fine detail recovery. 
Chen \etal~\cite{chen2020learning} construct an FSR network with face attention units that extend residual blocks with spatial attention branches.
%spatial attention residual networks for FSR. 
%by stacking face attention units each extending a vanilla residual block with a spatial attention branch. 
Jiang \etal~\cite{Jiang2022DualPathDF} propose an FSR network to capture and fuse both global facial shape and local facial components for the result. Dastmalchi \etal~\cite{dastmalchi2022super} propose to extract the wavelet coefficients and use them with the feature map in different scales to derive the result.

As a specific type of objects, faces have a well-defined structure. As such, face prior knowledge has been exploited by many FSR algorithms to guide the image generation. Song \etal~\cite{song2017learning} and Jiang \etal~\cite{jiang2018deep} both propose to crop a face into components by the estimated landmarks and predict high-frequency details for the components. 
Chen \etal~\cite{chen2018fsrnet} propose an NN that estimates both image features and landmark heatmaps/parsing maps and uses them to generate the final result. Li \etal~\cite{li2018face} propose a five-branch NN with each branch super-resolving one facial part. 
Yin \etal~\cite{yin2020joint} propose an NN that jointly conducts facial landmark detection and tiny face super-resolution. 
%Yin \etal~\cite{yin2020joint} propose a joint alignment and SR network that conducts facial landmark detection and tiny face super-resolution at the same time. 
Ma \etal~\cite{ma2020deep} propose an NN that conducts the FSR by iterative collaboration between facial image recovery and facial component estimation. Zhang \etal~\cite{zhang2020msfsr} 
%, Wu, and Chen~\cite{zhang2020msfsr} 
make a multi-stage NN that exploits enhanced facial boundaries for the FSR. %Kalarot, Li, and Porikli~\cite{kalarot2020component} 
Kalarot \etal~\cite{kalarot2020component} propose a multi-stage NN that makes facial component segmentation and patch cropping for the FSR. 
Li \etal~\cite{li2020learning} propose a two-stage framework and introduce face attributes and face boundaries information to promote the FSR.  
%Li \etal~\cite{li2020learning} propose a two-stage framework and introduce face attributes and face boundaries information to the two stages, respectively, to promote the FSR. 
Kim \etal~\cite{Kim2021EdgeAI} propose an NN that conducts FSR based on face edge information and identity loss function. Zhuang \etal~\cite{zhuang2022multi} propose an NN where an SR branch and a facial component estimation branch collaborate iteratively to generate the final result. Wang \etal~\cite{Wang2022PropagatingFP} propose to first learn face prior knowledge by a teacher FSR network and distill this knowledge to a student FSR network.

Note that the FSR algorithms reviewed above train an optimal model for a specific image up-scaling factor. Though some works (\eg~\cite{lu2020global,li2020learning,chen2020learning}) report results for multiple up-scaling factors, they train a separate model for each.

%In general, the FSR algorithms reviewed above train an optimal model for a specific image up-scaling factor. Though some works (\eg,~\cite{lu2020global,li2020learning,chen2020learning}) report experimental results for multiple up-scaling factors, they train a separate model for each. In other words, they do not make a generic model that can be trained once and applied to super-resolve face images of various resolutions.

\subsection{Scale-Arbitrary Image Super-Resolution}

%In addition, there are some SASR algorithms that can achieve SR for face images of varied low-resolutions by upscaling the input images with different scaling factors, although they are not specifically designed for FSR. 
For generic images, research has been started on scale-arbitrary super-resolution (SASR) in the recent few years. Hu \etal~\cite{hu2019meta} propose a meta-learning approach that dynamically predicts appropriate up-sampling filters for arbitrary magnification factors. Wang \etal~\cite{wang2021learning} design a plug-in module for existing SR networks which consists of multiple scale-aware filters and a scale-aware up-sampling layer. Chen \etal~\cite{chen2021learning} propose Local Implicit Image Function (LIIF) to predict the RGB value at any given image coordinate with the 2D deep features around the coordinate. %Because the image is represented as a continuous function, it can be mapping to any resolution. 
Son \etal~\cite{son2021srwarp} propose an SRWarp model that can transform the LR image to match any shape of the HR image at the feature level. 

Note that the SASR algorithms reviewed above are proposed for generic images but not optimized specifically for face images. In the field of FSR, we have found no similar algorithms published.

%\textcolor{red}{The algorithms mentioned above focus on image SR tasks and are not specially optimized for FSR. To the best of our knowledge, no scale-arbitrary algorithm has been proposed in the FSR field.}

\begin{figure*}[t]
\centering
\includegraphics[width=1.0\textwidth]{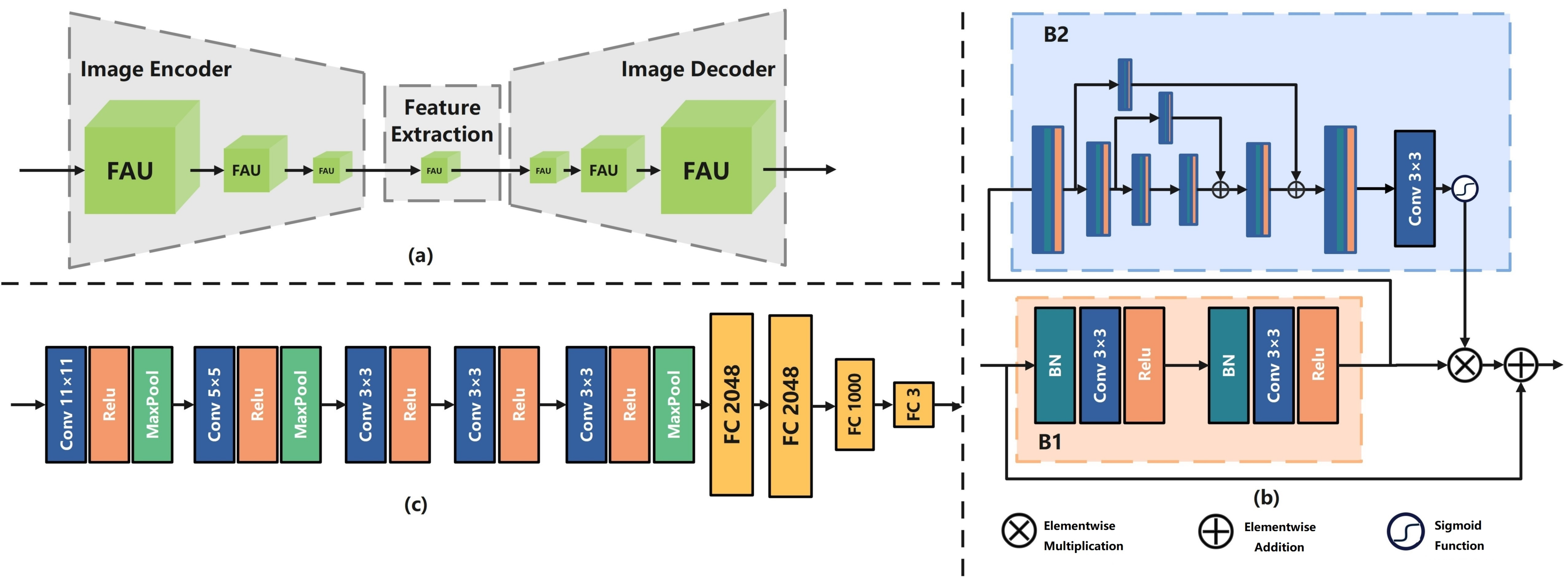} 
\caption{Structural overview of (a) anchor SR generator, (b) FAU block and (c) weight regressor. }
\label{fig:components}
\end{figure*}

\section{Approach} \label{sec:approach}

We propose a novel deep-learning-based unified framework that is trained for once and then used to super-resolve images of varied low resolutions. We name it UFVNet in a short form. The input image of an arbitrary resolution, $\mathbf{I}_{LR} \in \mathbb{R}^{h \times w}$, is firstly scaled by bi-cubic interpolation to $\mathbf{I}_{LS} \in \mathbb{R}^{H \times W}$ and then sent to the trained UFVNet to obtain the super-resolved image, $\mathbf{I}_{SR} \in \mathbb{R}^{H \times W}$, where $h \times w$ and $H \times W$ are the sizes of the LR image and the SR image, respectively. The architecture of the complete UFVNet is shown in Fig.~\ref{fig:architecture}.

As shown in Fig.~\ref{fig:architecture}, the complete UFVNet is composed of three encoder-decoder branches, a weight regression branch, a feature fusion module and a final decoder module. Note that, all the blocks in Fig.~\ref{fig:architecture} are used for the training, while only the green blocks are kept for the testing. The encoder-decoder in each branch itself is a complete FSR network that means to work best by itself for a certain input image resolution called anchor resolution. As such, the whole branch is an anchor SR generator (SRG) and the encoder/decoder is called anchor encoder/decoder. We train the network such that the three encoder-decoder branches correspond to three different anchor resolutions,
%, \eg $H/8 \times W/8$, $H/4 \times W/4$ and $H/2 \times W/2$, 
respectively, and the weight regressor automatically derives weights for the three encoded anchor features.

%assigns weights for the three encoded anchor features based on the closeness of the input image resolution to each anchor resolution.

Detailed explanations of the structural components, the training process and the loss function are given in the following subsections.

\subsection{Anchor SR Generator}
Theoretically, any encoder-decoder based FSR model may be used as backbone for the three anchor SR generators in UFVNet. Among them, SPARNet~\cite{chen2020learning} is of particular interest to us. Firstly, it is one of the state-of-the-art FSR models; secondly, it has a neatly defined encoder-decoder structure with no dependence on any face prior. Therefore, SPARNet fits our purpose of prototype neural network design very well.

In UFVNet, all the three anchor SR generators have the same network structure, as shown in Fig.~\ref{fig:components}(a). It is composed of an encoder with three face attention units (FAUs), a decoder also with three FAUs and one FAU in-between for further latent feature extraction. Note that we build the SR generator by adapting the original SPARNet~\cite{chen2020learning}. The major adaptations include that we adopt a light version (\ie SPARNet-Light-Attn3D~\cite{URL-SPARNet}) released by the authors %later 
and that we use just one FAU for the latent feature extraction favoring a compact and efficient network. As for the FAU, we use the same structure as released~\cite{URL-SPARNet}, which is shown in Fig.~\ref{fig:components}(b). It is composed of a feature extraction block and a spatial attention block plus a residual connection for effective high-level and low-level feature extraction~\cite{chen2020learning}.  

In essence, each anchor SR generator consists of seven FAUs in sequence. Denote the input and output of the $t$-th FAU as $\mathbf{f}_{t-1}$ and $\mathbf{f}_{t}$, $1 \leq t \leq 7$, respectively, with $\mathbf{f}_0 = \mathbf{I}_{LS}$. The computing process of the $t$-th FAU is mathematically modeled as:
\begin{eqnarray}
\mathbf{f}_t^{B_1} &=& B_1(\mathbf{f}_{t-1}), \\
\mathbf{f}_t^{B_2} &=& B_2(\mathbf{f}_t^{B_1}), \\
\mathbf{f}_t\ \ \ &=& \mathbf{f}_t^{B_1} \otimes \mathbf{f}_t^{B_2} \oplus \mathbf{f}_{t-1}
\label{eq:FAU}
\end{eqnarray}
where $B_1(\cdot)$ and $B_2(\cdot)$ are the mapping functions of the feature block and the attention block, respectively, and $\otimes$ and $\oplus$ denote element-wise multiplication and element-wise addition, respectively.

\begin{figure*}[t]
\centering
\includegraphics[width=1.0\textwidth]{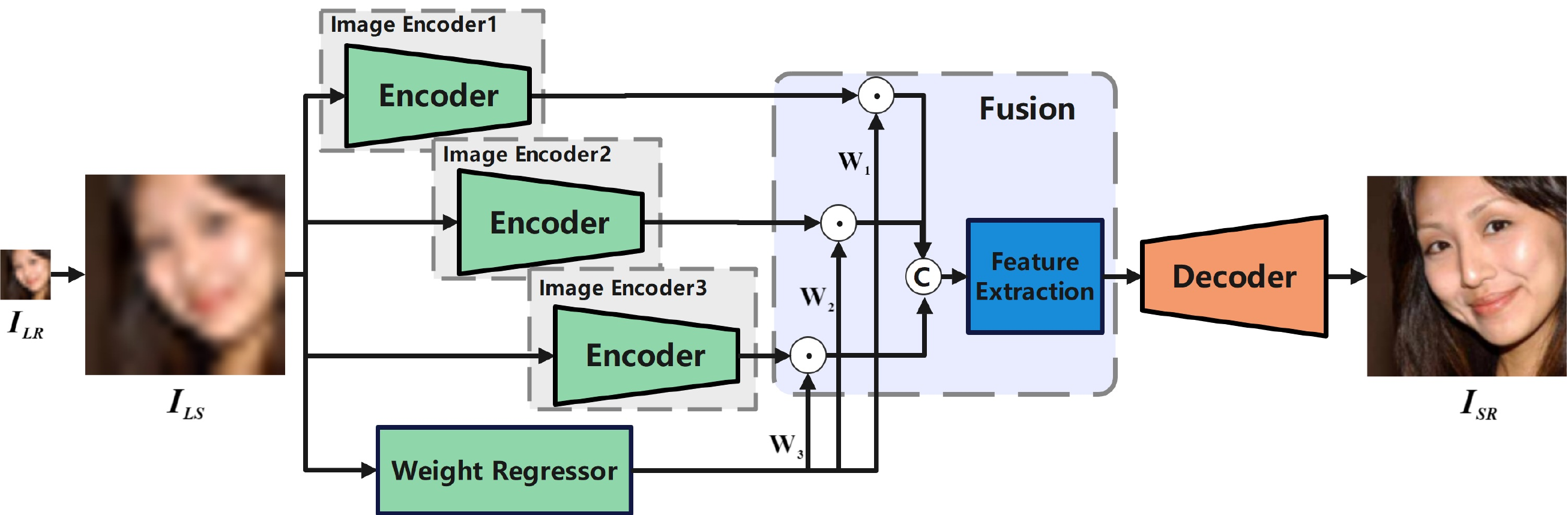} 
\caption{The portion of the complete UFVNet that is used for the integrated training. When trained, it is also the model that is finally used in testing.}
\label{fig:training}
\end{figure*}

\subsection{Weight Regressor}
Assume that we send the scaled input image $\mathbf{I}_{LS}$ to each anchor SR generator separately, we would obtain three SR images. It is predictable that the best result would be mostly given by the anchor SR generator whose anchor resolution is the closest to the resolution of $\mathbf{I}_{LR}$. However, the resolution of $\mathbf{I}_{LR}$ may not exactly match any one of the anchor resolutions. As such, we send $\mathbf{I}_{LS}$ (scaled $\mathbf{I}_{LR}$) to all the three anchor encoders and make a weighted fusion of the encoded features to make a more robust and complete characterization of $\mathbf{I}_{LS}$. While a simple formula may be manually defined to map input resolution to anchor weights, we choose to design a neural network module for weight regression that can learn a more complex mapping adaptive to both resolution and content of $\mathbf{I}_{LR}$.

All the weights are non-negative and sum up to one. %In general, we should assign bigger weights to anchor resolutions closer to the input resolution. 
In general, we should assign a bigger weight to an anchor resolution closer to the input resolution. If we view the three anchor resolutions as cluster heads and weights as confidence values, weight computing turns out to be similar to soft classification of the input image by resolution. Therefore, we employ AlexNet, a concise image classifier, as the basis of our weight regressor. Upon the baseline, we modify the hyper-parameters of the first maxpooling layer to fit the size of $\mathbf{I}_{LS}$,
%the resolution of $\mathbf{I}_{LS}$ (or $\mathbf{I}_{SR}$), 
and add a fully-connected layer of three neurons at the end. The structure of the weight regressor is shown in Fig.~\ref{fig:components}(c).

\subsection{Final Fusion and Decoding}

At the end, the anchor features produced by the three anchor encoders are fused and decoded to generate the final SR image.  As shown in Fig.~\ref{fig:architecture}, the fusion is accomplished by a weighted concatenation of the three encoded anchor features followed by another feature extraction block that is structurally the same as that in Fig.~\ref{fig:components}(a).  Finally, the decoder at the end of the UFVNet also has the same structure as that in Fig.~\ref{fig:components}(a).

\subsection{Training Process}

All the blocks and data flows as shown in Fig.~\ref{fig:architecture} participate in the training of UFVNet, and the training is conducted in two stages. Firstly, all the anchor SR generators and the weight regressor are separately trained; secondly, the modules pre-trained in the first stage are integrated and a portion of the UFVNet is trained further as a whole. For convenience, we also call the two stages separate training and integrated training, respectively.

We define the three anchor resolutions to be $\frac{1}{8}$, $\frac{1}{4}$ and $\frac{1}{2}$ the HR image size and train the anchor SR generators correspondingly. In order to generate the training LR images, we down-sample each HR image in the training set to $\frac{1}{8}$, $\frac{1}{4}$ and $\frac{1}{2}$ the original size via bi-cubic interpolation. For the three anchor SR generators, we use the $\frac{1}{8}$, $\frac{1}{4}$ and $\frac{1}{2}$ down-sampled LR images, respectively, and the corresponding HR images for the training. Each anchor SR generator is then trained as a stand-alone FSR model.

In order to train the weight regressor, we down-sample each HR image in the training set to eight low resolutions (\ie $\frac{1}{16}$, $\frac{2}{16}$, $\frac{3}{16}$, $\frac{4}{16}$, $\frac{5}{16}$, $\frac{6}{16}$, $\frac{7}{16}$, $\frac{8}{16}$ the original HR image size) via bi-cubic interpolation. The weight regressor is then trained as a soft classifier. Denote the down-sampling rates of the three anchor resolutions as $a_1$, $a_2$ and $a_3$. Given an LR image with down-sampling rate $r$, its class score $W = (w_1, w_2, w_3)$ is defined by
\begin{eqnarray}
d_i &=& \left| r-a_i \right|, 1 \leq i \leq 3, \nonumber \\
w_i &=& \frac{1}{2} \times (1 - \frac{d_i}{d_1+d_2+d_3}), 1 \leq i \leq 3.
\label{eqn:weight}
\end{eqnarray}

Finally, we integrate all the separately trained modules and further train a portion of the UFVNet as a whole. Note that, in this stage, the feature extraction and the decoder blocks of the anchor SR generators are discarded and the remaining portion of the UFVNet, as shown in Fig.~\ref{fig:training}, is trained by forward and backward propagations. After training, the neural network shown in Fig.~\ref{fig:training} is then the final model used for testing.  

In the following, we only formulate the forward propagation, from which the backward propagation may be derived. Denote the functions corresponding to the weight regressor, the three anchor encoders, the feature extraction block and the final decoder as $F_{wt}$, $F_{enc1}$, $F_{enc2}$, $F_{enc3}$, $F_{FE}$ and $F_{dec}$, respectively. The whole mapping from the input $\mathbf{I}_{LR}$ to the output $\mathbf{I}_{SR}$ is formulated as
\begin{eqnarray}
W\qquad &=& F_{wt} ( U(\mathbf{I}_{LR}) ),\\
\mathbf{f}_{enc1}\quad &=& F_{enc1}(U(\mathbf{I}_{LR})) \cdot W[1],\\
\mathbf{f}_{enc2}\quad &=& F_{enc2}(U(\mathbf{I}_{LR})) \cdot W[2],\\
\mathbf{f}_{enc3}\quad &=& F_{enc3}(U(\mathbf{I}_{LR})) \cdot W[3],\\
\mathbf{f}_{concat}\: &=& C( \mathbf{f}_{enc1}, \mathbf{f}_{enc2}, \mathbf{f}_{enc3}),\\
\mathbf{I}_{SR}\quad\; &=& F_{dec}(F_{FE}(\mathbf{f}_{concat}))
\end{eqnarray}
where $W[i], 1 \leq i \leq 3$, denotes the $i$-th weight, $U(\cdot)$ denotes the up-scaling operation, and $C(\cdot)$ denotes the concatenation operation.

\subsection{Loss Functions}

%, and each $\mathbf{I}_{LR}^{i}$ is scaled to $\mathbf{I}_{LS}^{i}$ before it is fed to the network.

In the separate training stage, $L_1$ loss is used for each SR generator and cross-entropy loss is used for the weight regressor. 
Denote the training LR-HR image pairs as ${\{\mathbf{I}_{LR}^{i},\mathbf{I}_{HR}^{i}\}}_{i=1}^{N}$. For an SR generator (and the whole FSR neural network in Fig.~\ref{fig:training}), the loss is defined as 
\begin{equation}
    L^{SR} = \frac{1}{N}\sum_{i=0}^{N}\left| \mathbf{I}_{HR}^{i}-\mathbf{I}_{SR}^{i} \right|.
\label{eq:lossimage}
\end{equation}
For the weight regressor, the loss is defined as
\begin{equation}
     L^{wt} = \frac{1}{N}\sum_{i=0}^{N}L_{CE}({W}_{GT}^{i}-{W}^{i})
\label{eq:losswt}
\end{equation}
where ${W}^{i}$ is the estimated weight vector and ${W}_{GT}^{i}$ is the ground-truth weight vector (computed by Eq.~\ref{eqn:weight}) for $\mathbf{I}_{LR}^{i}$, and $L_{CE}(\cdot)$ denotes the cross-entropy loss function.

In the integrated training stage, the neural network shown in Fig.~\ref{fig:training} is trained as a whole. The total loss is defined as
\begin{equation}
     L^{total} = L^{SR}+ \alpha L^{wt}
\end{equation}
with $L^{SR}$ and $L^{wt}$ defined in Eq.s~\ref{eq:lossimage} and \ref{eq:losswt}, respectively. We empirically set $\alpha =0.1$ in the experiments.

\begin{figure*}[htbp]
\centering

    {\rotatebox{90}{\scriptsize{~~~~~$8\times8$}}}
    \begin{minipage}[t]{0.12\linewidth}
        \centering
        \includegraphics[width=\textwidth]{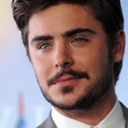}
    \end{minipage}%
    \begin{minipage}[t]{0.12\linewidth}
        \centering
        \includegraphics[width=\textwidth]{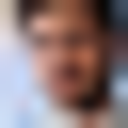}
    \end{minipage}%
    \begin{minipage}[t]{0.12\linewidth}
        \centering
        \includegraphics[width=\textwidth]{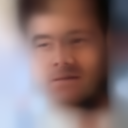}
    \end{minipage}%
    \begin{minipage}[t]{0.12\linewidth}
        \centering
        \includegraphics[width=\textwidth]{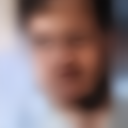}
    \end{minipage}%
    \begin{minipage}[t]{0.12\linewidth}
        \centering
        \includegraphics[width=\textwidth]{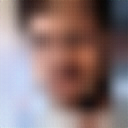}
    \end{minipage}%
    \begin{minipage}[t]{0.12\linewidth}
        \centering
       \includegraphics[width=\textwidth]{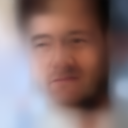}
    \end{minipage}%
    \begin{minipage}[t]{0.12\linewidth}
        \centering
        \includegraphics[width=\textwidth]{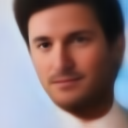}
    \end{minipage}
    
    {\rotatebox{90}{\scriptsize{~~~~~$8\times8$}}}
    \begin{minipage}[t]{0.12\linewidth}
        \centering
        \includegraphics[width=\textwidth]{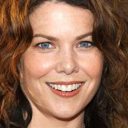}
    \end{minipage}%
    \begin{minipage}[t]{0.12\linewidth}
        \centering
        \includegraphics[width=\textwidth]{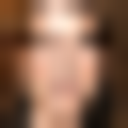}
    \end{minipage}%
    \begin{minipage}[t]{0.12\linewidth}
        \centering
        \includegraphics[width=\textwidth]{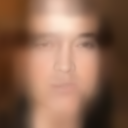}
    \end{minipage}%
    \begin{minipage}[t]{0.12\linewidth}
        \centering
        \includegraphics[width=\textwidth]{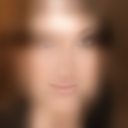}
    \end{minipage}%
    \begin{minipage}[t]{0.12\linewidth}
        \centering
        \includegraphics[width=\textwidth]{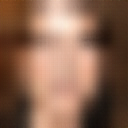}
    \end{minipage}%
    \begin{minipage}[t]{0.12\linewidth}
        \centering
        \includegraphics[width=\textwidth]{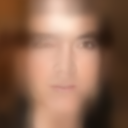}
    \end{minipage}%
    \begin{minipage}[t]{0.12\linewidth}
        \centering
        \includegraphics[width=\textwidth]{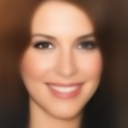}
    \end{minipage}
    \hspace{-1mm}
    
    {\rotatebox{90}{\scriptsize{~~~~~$24\times24$}}}
    \begin{minipage}[t]{0.12\linewidth}
        \centering
        \includegraphics[width=\textwidth]{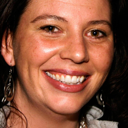}
    \end{minipage}%
    \begin{minipage}[t]{0.12\linewidth}
        \centering
        \includegraphics[width=\textwidth]{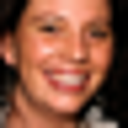}
    \end{minipage}%
    \begin{minipage}[t]{0.12\linewidth}
        \centering
        \includegraphics[width=\textwidth]{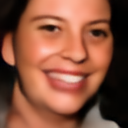}
    \end{minipage}%
    \begin{minipage}[t]{0.12\linewidth}
        \centering
        \includegraphics[width=\textwidth]{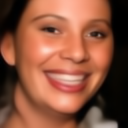}
    \end{minipage}%
    \begin{minipage}[t]{0.12\linewidth}
        \centering
        \includegraphics[width=\textwidth]{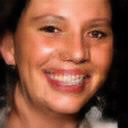}
    \end{minipage}%
    \begin{minipage}[t]{0.12\linewidth}
        \centering
        \includegraphics[width=\textwidth]{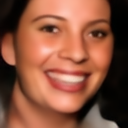}
    \end{minipage}%
    \begin{minipage}[t]{0.12\linewidth}
        \centering
        \includegraphics[width=\textwidth]{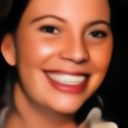}
    \end{minipage}
    
    {\rotatebox{90}{\scriptsize{~~~~~$24\times24$}}}
     \begin{minipage}[t]{0.12\linewidth}
        \centering
        \includegraphics[width=\textwidth]{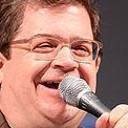}
    \end{minipage}%
    \begin{minipage}[t]{0.12\linewidth}
        \centering
       \includegraphics[width=\textwidth]{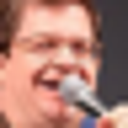}
    \end{minipage}%
    \begin{minipage}[t]{0.12\linewidth}
        \centering
        \includegraphics[width=\textwidth]{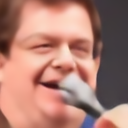}
    \end{minipage}%
    \begin{minipage}[t]{0.12\linewidth}
        \centering
        \includegraphics[width=\textwidth]{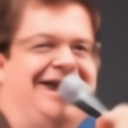}
     \end{minipage}%
     \begin{minipage}[t]{0.12\linewidth}
        \centering
        \includegraphics[width=\textwidth]{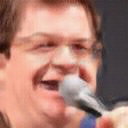}
    \end{minipage}%
    \begin{minipage}[t]{0.12\linewidth}
        \centering
        \includegraphics[width=\textwidth]{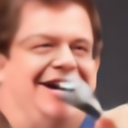}
    \end{minipage}%
    \begin{minipage}[t]{0.12\linewidth}
        \centering
        \includegraphics[width=\textwidth]{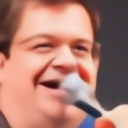}
    \end{minipage}
    
    {\rotatebox{90}{\scriptsize{~~~~~$56\times56$}}}
    \begin{minipage}[t]{0.12\linewidth}
        \centering
        \includegraphics[width=\textwidth]{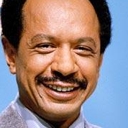}
        \centerline{\scriptsize{HR}}
    \end{minipage}%
    \begin{minipage}[t]{0.12\linewidth}
        \centering
        \includegraphics[width=\textwidth]{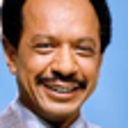}
        \centerline{\scriptsize{Bicubic}}
    \end{minipage}%
    \begin{minipage}[t]{0.12\linewidth}
        \centering
        \includegraphics[width=\textwidth]{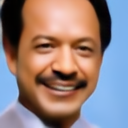}
        \centerline{\scriptsize{DIC}}
    \end{minipage}%
    \begin{minipage}[t]{0.12\linewidth}
        \centering
        \includegraphics[width=\textwidth]{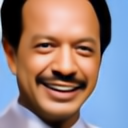}
        \centerline{\scriptsize{SISN}}
    \end{minipage}%
    \begin{minipage}[t]{0.12\linewidth}
        \centering
        \includegraphics[width=\textwidth]{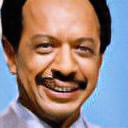}
        \centerline{\scriptsize{WIPA}}
    \end{minipage}%
    \begin{minipage}[t]{0.12\linewidth}
        \centering
        \includegraphics[width=\textwidth]{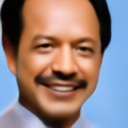}
        \centerline{\scriptsize{MLGDN}}
    \end{minipage}%
    \begin{minipage}[t]{0.12\linewidth}
        \centering
        \includegraphics[width=\textwidth]{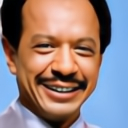}
        \centerline{\scriptsize{UFVNet}}
    \end{minipage}
    \hspace{-2mm}

   \vspace{0.2cm}

   \caption{Visual results by UFVNet and the benchmark FR methods on example images from CelebA-MR8 and Helen-MR8.}
\label{fig:fr}
\end{figure*}

\section{Experiments} 

\subsection{Implementation Details}

\textbf{Datasets and metrics:} The datasets for training and testing in our experiments are variant from two prevalently
%widely 
used datasets: CelebA~\cite{liu2015deep} and Helen~\cite{le2012interactive}. 
We preprocess the images in the two datasets following the settings in Ma \etal~\cite{ma2020deep}. Specifically, we crop the face images in the datasets with the files containing 68 key point position information of each face provided by~\cite{ma2020deep}, and resize the cropped images %them 
to $128 \times 128$ to obtain the %\pjl{reference} %original 
HR images. We then scale the HR images to different low resolutions via bi-cubic interpolation and build up five multi-resolution data sets called CelebA-MR3, CelebA-MR7, CelebA-MR8, Helen-MR7 and Helen-MR8, respectively. 
%The scaled face images in each variant dataset have a one-to-one correspondence to the images in the original dataset.
There are three different low resolutions (\ie $\frac{1}{8}$,  $\frac{1}{4}$, $\frac{1}{2}$ of the original HR) in CelebA-MR3, seven (\ie $\frac{2}{16}$, $\frac{3}{16}$, $\frac{4}{16}$, $\frac{5}{16}$, $\frac{6}{16}$, $\frac{7}{16}$, $\frac{8}{16}$ of the original HR) in CelebA-MR7 and Helen-MR7, and eight (\ie $\frac{1}{16}$, $\frac{2}{16}$, $\frac{3}{16}$, $\frac{4}{16}$, $\frac{5}{16}$, $\frac{6}{16}$, $\frac{7}{16}$, $\frac{8}{16}$ of the original HR) in CelebA-MR8 and Helen-MR8.
%There are 8 different low resolutions (\ie $\frac{1}{16}$, $\frac{2}{16}$, $\frac{3}{16}$, $\frac{4}{16}$, $\frac{5}{16}$, $\frac{6}{16}$, $\frac{7}{16}$, $\frac{8}{16}$ of the original HR) in CelebA-MR8 and Helen-MR8, and 3 (\ie $\frac{1}{2}$,  $\frac{1}{4}$, $\frac{1}{8}$ of the original HR) in CelebA-MR3.
In each of these datasets, there is a balanced distribution of images among various resolutions.
We also follow the same settings in~\cite{ma2020deep,zhu2016deep,chen2020learning} to divide the five variant datasets into training sets and testing sets. In all our experiments, we use the training set of CelebA-MR3, CelebA-MR7 or CelebA-MR8 for training and use the testing set of CelebA-MR7, CelebA-MR8, Helen-MR7 or Helen-MR8 for testing. We use PSNR and SSIM ~\cite{wang2004image} to evaluate the quality of an SR image.

\textbf{Training settings:} For UFVNet, we set the batch size to 32, and fix the learning rate at $2\times10^{-4}$. We use Adam to optimize the model with $\beta_{1}=0.9$, $\beta_{2} = 0.999 $. Our models are implemented in PyTorch and run on NVIDIA RTX 2080Ti GPUs.

\begin{table}[b]
\caption{PSNR and SSIM performance of UFVNet and the benchmark methods using fixed-resolution input. In each column, result in \textbf{bold} is the best. }\label{tab:fr}
\vspace{1mm}
\resizebox{.95\columnwidth}{!}{
\begin{tabular}{|c|c|c|c|c|c|}
\hline
 \multirow{2}{*}{\diagbox{Method}{Dataset}} & Scale & \multicolumn{2}{c|}{CelebA-MR8} & \multicolumn{2}{c|}{Helen-MR8}\\
\cline{3-6}
 & Factor
 & \hspace{-2mm} PSNR$\uparrow$ \hspace{-3mm}
 & \hspace{-1mm} SSIM$\uparrow$ \hspace{-3mm} 
 & \hspace{-1mm} PSNR$\uparrow$ \hspace{-3mm} 
 & \hspace{-1mm} SSIM$\uparrow$ \hspace{-1mm}\\
\hline
\hline
\hspace{-2mm} Bicubic \hspace{-2mm}&- & 27.71 & 0.7853&28.21 & 0.8238 \\
\hline
%\hspace{-2mm} Bicubic* \hspace{-2mm}&-&25.74 & 0.7541 & 27.23 & 0.7083 \\
%\hline
\hspace{-2mm} DIC ~\cite{ma2020deep}\hspace{-2mm}& 8$\times$ &26.50 & 0.7604 & 25.99 & 0.7654 \\
\hline
\hspace{-2mm} SISN ~\cite{lu2021face}\hspace{-2mm} & 4$\times$ & 27.45 & 0.7763 & 27.63 & 0.8067 \\
\hline
\hspace{-2mm} WIPA ~\cite{dastmalchi2022super} \hspace{-2mm}& 4$\times$ & 26.75 & 0.7624 & 27.11 & 0.7994 \\
\hline
\hspace{-2mm} MLGDN~\cite{zhuang2022multi} \hspace{-2mm}& 8$\times$ & 26.39 & 0.7569 &25.76 & 0.7565  \\
\hline
\hspace{-2mm} UFVNet(Ours) \hspace{-2mm} &-& \textbf{29.46} & \textbf{0.8321}&\textbf{29.49} & \textbf{0.8446 } \\
\hline
\end{tabular}}
\end{table}

\begin{figure}[htbp]
\centering
    %{\rotatebox{90}{\scriptsize{~~~~~$16\times16$}}}
    %\begin{minipage}[t]{0.19\linewidth}
    %    \centering
    %    \includegraphics[width=\textwidth]{visualimage/image1/HR/201722.png}
    %\end{minipage}%
    %\begin{minipage}[t]{0.19\linewidth}
    %    \centering
    %    \includegraphics[width=\textwidth]{visualimage/image1/BIC/BIC16.png}
    %\end{minipage}%
    %\begin{minipage}[t]{0.19\linewidth}
    %    \centering
    %    \includegraphics[width=\textwidth]{visualimage/image1/DIC/DIC16.png}
    %\end{minipage}%
    %\begin{minipage}[t]{0.19\linewidth}
    %    \centering
    %    \includegraphics[width=\textwidth]{visualimage/image1/WIPA/WIPA16.png}
    %\end{minipage}%
    %\begin{minipage}[t]{0.19\linewidth}
    %    \centering
    %    \includegraphics[width=\textwidth]{visualimage/image1/UFVNet/UFV16.png}
    %\end{minipage}
    {\rotatebox{90}{\scriptsize{~~~~~$16\times16$}}}
    \begin{minipage}[t]{0.19\linewidth}
        \centering
        \includegraphics[width=\textwidth]{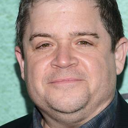}
    \end{minipage}%
    \begin{minipage}[t]{0.19\linewidth}
        \centering
        \includegraphics[width=\textwidth]{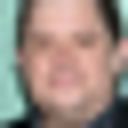}
    \end{minipage}%
    \begin{minipage}[t]{0.19\linewidth}
        \centering
        \includegraphics[width=\textwidth]{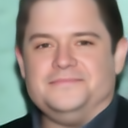}
    \end{minipage}%
    \begin{minipage}[t]{0.19\linewidth}
        \centering
        \includegraphics[width=\textwidth]{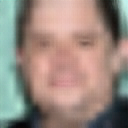}
    \end{minipage}%
    \begin{minipage}[t]{0.19\linewidth}
        \centering
        \includegraphics[width=\textwidth]{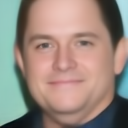}
    \end{minipage}
    \hspace{1mm}
   {\rotatebox{90}{\scriptsize{~~~~~$32\times32$}}}
    \begin{minipage}[t]{0.19\linewidth}
        \centering
        \includegraphics[width=\textwidth]{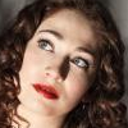}
        \centerline{\scriptsize{HR}}
    \end{minipage}%
    \begin{minipage}[t]{0.19\linewidth}
        \centering
        \includegraphics[width=\textwidth]{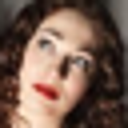}
        \centerline{\scriptsize{Bicubic}}
    \end{minipage}%
    \begin{minipage}[t]{0.19\linewidth}
        \centering
        \includegraphics[width=\textwidth]{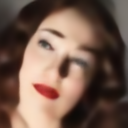}
        \centerline{\scriptsize{DIC}}
    \end{minipage}%
    \begin{minipage}[t]{0.19\linewidth}
        \centering
        \includegraphics[width=\textwidth]{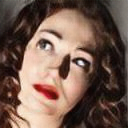}
        \centerline{\scriptsize{WIPA}}
    \end{minipage}%
    \begin{minipage}[t]{0.19\linewidth}
        \centering
        \includegraphics[width=\textwidth]{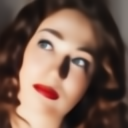}
        \centerline{\scriptsize{UFVNet}}
    \end{minipage}
    \hspace{1mm}
    %{\rotatebox{90}{\scriptsize{~~~~~$32\times32$}}}
    %\begin{minipage}[t]{0.19\linewidth}
    %    \centering
    %    \includegraphics[width=\textwidth]{visualimage/image1/HR/201578.png}
    %    \centerline{HR}
    %\end{minipage}%
    %\begin{minipage}[t]{0.19\linewidth}
    %    \centering
    %    \includegraphics[width=\textwidth]{visualimage/image1/BIC/201578.png}
    %    \centerline{Bicubic}
    %\end{minipage}%
    %\begin{minipage}[t]{0.19\linewidth}
    %    \centering
    %    \includegraphics[width=\textwidth]{visualimage/image1/DIC/201578.png}
    %    \centerline{DIC}
    %\end{minipage}%
    %\begin{minipage}[t]{0.19\linewidth}
    %    \centering
    %    \includegraphics[width=\textwidth]{visualimage/image1/WIPA/201578.png}
    %    \centerline{WIPA}
    %\end{minipage}%
    %\begin{minipage}[t]{0.19\linewidth}
    %    \centering
    %    \includegraphics[width=\textwidth]{visualimage/image1/UFVNet/201578.png}
    %    \centerline{UFVNet}
    %\end{minipage}
    %\caption{A visual comparison on the state-of-the-art methods using fixed-resolution input and UFVNet. The input resolution of first rows ($16\times16$) is that of DIC (8$\times$), and the input resolution of last rows ($32\times32$) is that of WIPA (4$\times$). The fixed-resolution methods shows better performance when the inputs match the resolution it is specifically trained for.}
%    \caption{A visual comparison on the state-of-the-art methods using fixed-resolution input and UFVNet. }
    \vspace{0.1cm}
    \caption{More visual comparison for Bicubic, DIC, WIPA and UFVNet. DIC is originally trained for $16 \times 16$ input while WIPA for $32 \times 32$ input. }
\label{fig:or_celeba_exp3}
\end{figure}

\subsection{Comparison with the state-of-the-arts}
The previously published FSR methods cannot natively support generating SR images from inputs of varied low resolutions with a single model.
%Some of the methods scale LR images to the original resolution before inputting, but their model can not solve the input well if we merely alter their testing set. Others are desinged for fixed up-scaling proportion and use face images at fixed resolution as the input.
Some of the SOTA methods are designed for a fixed up-scaling factor and use face images at a fixed resolution as the input. The others scale an LR image of arbitrary size uniformly to the original resolution (of the HR image) before feeding it to the network and, as such, may theoretically super-resolve input images of various resolutions. But still, the authors usually train and test their models for specific fixed input resolutions.
%but their models fail at majority of the inputs if we merely alter their testing set. 
%For a complete evaluation, we conduct two specialized experiments to compare UFVNet with the methods using fixed-resolution (FR) input and the methods using original-resolution (OR) input, respectively. For convenience of description, they may be abbreviated as FR methods and OR methods, respectively. 
%\textcolor{red}{To evaluate UFVNet comprehensively, we conduct three specific experiments, comparing it with FSR methods that use fixed-resolution (FR) input and original-resolution (OR) input, as well as image SR methods that use scale-arbitrary input. For brevity, we refer to them as FR methods, OR methods, and SASR methods, respectively.}
For comprehensive evaluation, we conduct experiments to compare UFVNet with both FSR methods using fixed-resolution (FR) input and FSR methods using original-resolution (OR) input. Further, we compare UFVNet with SASR models that are originally proposed for generic images. For brevity, we refer to these benchmark methods as FR methods, OR methods and SASR methods, respectively.

%\textcolor{red}{For a complete evaluation, we conduct three specialized experiments to compare UFVNet with the methods using fixed-resolution (FR) input, the methods using original-resolution (OR) input, respectively and scale-arbitrary super-resolution(SASR) methods in image super-resolution. For convenience of description, they may be abbreviated as FR methods, OR methods and SASR methods, respectively.}

%We compare our proposed UFVNet method with state-of-the-art methods, it is important to note that existing face super-resolution networks can be divided into two broad categories.The first type of network is to use the LR image to obtain a scaled low-resolution image at the beginning of the network architecture (by upsampling the LR image to the network's SR image size using bicubic interpolation), and then use the scaled low-resolution image as the input of the network; the other is to directly take the LR image as the input of the network.

\begin{table*}\centering
\caption{PSNR and SSIM performance of UFVNet and the benchmark methods using original-resolution input. In each column, result in \textbf{bold} is the best. 
}\label{tab:or}
\vspace{0.1cm}
\begin{tabular}{|c|c|c|c|c|c|c|c|c|}
\hline
 \multirow{3}{*}{\diagbox{Method}{Dataset}}
 &\multicolumn{4}{c|}{Trained on CelebA-MR8}&\multicolumn{4}{c|}{Trained on CelebA-MR3}\\
 \cline{2-9}
 & \multicolumn{2}{c|}{CelebA-MR8}& \multicolumn{2}{c|}{Helen-MR8}  &\multicolumn{2}{c|}{CelebA-MR8} & \multicolumn{2}{c|}{Helen-MR8}\\
\cline{2-9}
 & \hspace{-2mm} PSNR$\uparrow$ \hspace{-3mm}
 & \hspace{-3mm} SSIM$\uparrow$ \hspace{-3mm} 
 & \hspace{-3mm} PSNR$\uparrow$ \hspace{-3mm} 
 & \hspace{-1mm} SSIM$\uparrow$ \hspace{-1mm}
 & \hspace{-2mm} PSNR$\uparrow$ \hspace{-3mm}
 & \hspace{-3mm} SSIM$\uparrow$ \hspace{-3mm} 
 & \hspace{-3mm} PSNR$\uparrow$ \hspace{-3mm} 
 & \hspace{-1mm} SSIM$\uparrow$ \hspace{-1mm}\\
\hline
\hline
\hspace{-2mm} Bicubic \hspace{-2mm} & 27.71 & 0.7853&28.21 & 0.8238 & 27.71 & 0.7853&28.21 & 0.8238\\
\hline
\hspace{-2mm} SPARNet~\cite{chen2020learning} \hspace{-2mm}& 29.20&0.8201 & 29.34 & 0.8406 & 28.73& 0.8112& 28.90& 0.8325\\
\hline
\hspace{-2mm} EIPNet~\cite{Kim2021EdgeAI}\hspace{-2mm} & 25.11 & 0.7152 & 24.84 & 0.7303 & 24.89 & 0.7073 & 24.59 & 0.7243\\
\hline
\hspace{-2mm} KDFSRNet~\cite{Wang2022PropagatingFP}\hspace{-2mm}&26.49 & 0.7672 & 26.83 & 0.8041 &26.49 & 0.7672 & 26.83 & 0.8041  \\
\hline
\hspace{-2mm} UFVNet(Ours)  & \textbf{29.46} & \textbf{0.8321}&\textbf{29.49} & \textbf{0.8446} &\textbf{29.16}&\textbf{0.8241} &\textbf{29.46}&\textbf{0.8453}\\
\hline
\end{tabular}
\end{table*}

\begin{figure*}[htbp]

\centering
    {\rotatebox{90}{\scriptsize{~~~~~$16\times16$}}}
    \begin{minipage}[t]{0.09\linewidth}
        \centering
        \includegraphics[width=\textwidth]{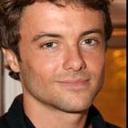}
    \end{minipage}%
    \begin{minipage}[t]{0.09\linewidth}
        \centering
        \includegraphics[width=\textwidth]{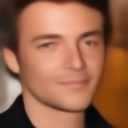}
    \end{minipage}%
    \begin{minipage}[t]{0.09\linewidth}
        \centering
        \includegraphics[width=\textwidth]{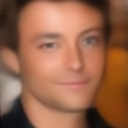}
    \end{minipage}%
    \begin{minipage}[t]{0.09\linewidth}
        \centering
        \includegraphics[width=\textwidth]{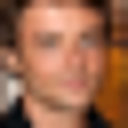}
    \end{minipage}%
    \begin{minipage}[t]{0.09\linewidth}
        \centering
        \includegraphics[width=\textwidth]{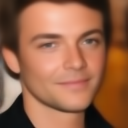}
    \end{minipage}%
    \hspace{1mm}
    {\rotatebox{90}{\scriptsize{~~~~~~~$8\times8$}}}
    \begin{minipage}[t]{0.09\linewidth}
        \centering
        \includegraphics[width=\textwidth]{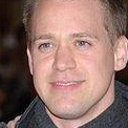}
    \end{minipage}%
    \begin{minipage}[t]{0.09\linewidth}
        \centering
        \includegraphics[width=\textwidth]{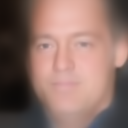}
    \end{minipage}%
    \begin{minipage}[t]{0.09\linewidth}
        \centering
        \includegraphics[width=\textwidth]{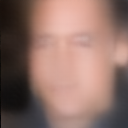}
    \end{minipage}%
    \begin{minipage}[t]{0.09\linewidth}
        \centering
        \includegraphics[width=\textwidth]{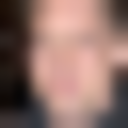}
    \end{minipage}%
    \begin{minipage}[t]{0.09\linewidth}
        \centering
        \includegraphics[width=\textwidth]{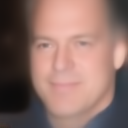}
    \end{minipage}
    
    {\rotatebox{90}{\scriptsize{~~~~~$32\times32$}}}
    \begin{minipage}[t]{0.09\linewidth}
        \centering
        \includegraphics[width=\textwidth]{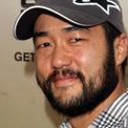}
    \end{minipage}%
    \begin{minipage}[t]{0.09\linewidth}
        \centering
        \includegraphics[width=\textwidth]{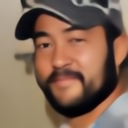}
    \end{minipage}%
    \begin{minipage}[t]{0.09\linewidth}
        \centering
        \includegraphics[width=\textwidth]{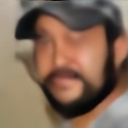}
    \end{minipage}%
    \begin{minipage}[t]{0.09\linewidth}
        \centering
        \includegraphics[width=\textwidth]{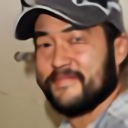}
    \end{minipage}%
    \begin{minipage}[t]{0.09\linewidth}
        \centering
        \includegraphics[width=\textwidth]{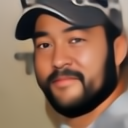}
    \end{minipage}%
    \hspace{1mm}
    {\rotatebox{90}{\scriptsize{~~~~~$24\times24$}}}
    \begin{minipage}[t]{0.09\linewidth}
        \centering
        \includegraphics[width=\textwidth]{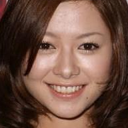}
    \end{minipage}%
    \begin{minipage}[t]{0.09\linewidth}
        \centering
        \includegraphics[width=\textwidth]{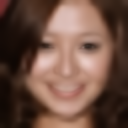}
    \end{minipage}%
    \begin{minipage}[t]{0.09\linewidth}
        \centering
        \includegraphics[width=\textwidth]{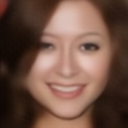}
    \end{minipage}%
    \begin{minipage}[t]{0.09\linewidth}
        \centering
        \includegraphics[width=\textwidth]{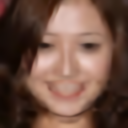}
    \end{minipage}%
    \begin{minipage}[t]{0.09\linewidth}
        \centering
        \includegraphics[width=\textwidth]{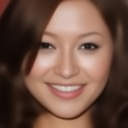}
    \end{minipage}

    {\rotatebox{90}{\scriptsize{~~~~~$40\times40$}}}
    \begin{minipage}[t]{0.09\linewidth}
        \centering
        \includegraphics[width=\textwidth]{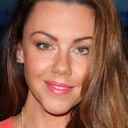}
    \end{minipage}%
    \begin{minipage}[t]{0.09\linewidth}
        \centering
        \includegraphics[width=\textwidth]{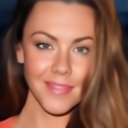}
    \end{minipage}%
    \begin{minipage}[t]{0.09\linewidth}
        \centering
        \includegraphics[width=\textwidth]{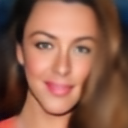}
    \end{minipage}%
    \begin{minipage}[t]{0.09\linewidth}
        \centering
    \includegraphics[width=\textwidth]{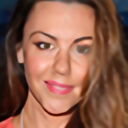}
    \end{minipage}%
    \begin{minipage}[t]{0.09\linewidth}
        \centering
        \includegraphics[width=\textwidth]{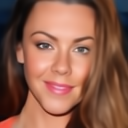}
    \end{minipage}%
    \hspace{1mm}
    {\rotatebox{90}{\scriptsize{~~~~~$40\times40$}}}
    \begin{minipage}[t]{0.09\linewidth}
        \centering
        \includegraphics[width=\textwidth]{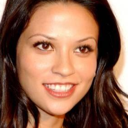}
    \end{minipage}%
    \begin{minipage}[t]{0.09\linewidth}
        \centering
        \includegraphics[width=\textwidth]{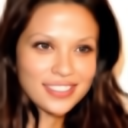}
    \end{minipage}%
    \begin{minipage}[t]{0.09\linewidth}
        \centering
        \includegraphics[width=\textwidth]{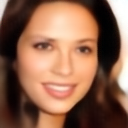}
    \end{minipage}%
    \begin{minipage}[t]{0.09\linewidth}
        \centering
        \includegraphics[width=\textwidth]{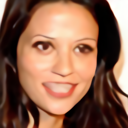}
    \end{minipage}%
    \begin{minipage}[t]{0.09\linewidth}
        \centering
        \includegraphics[width=\textwidth]{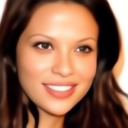}
    \end{minipage}

    {\rotatebox{90}{\scriptsize{~~~~~$64\times64$}}}
    \begin{minipage}[t]{0.09\linewidth}
        \centering
        \includegraphics[width=\textwidth]{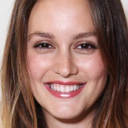}
        \centerline{\scriptsize{HR}}
    \end{minipage}%
    \begin{minipage}[t]{0.09\linewidth}
        \centering
        \includegraphics[width=\textwidth]{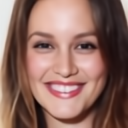}
        \centerline{\scriptsize{SPARNet}}
    \end{minipage}%
    \begin{minipage}[t]{0.09\linewidth}
        \centering
        \includegraphics[width=\textwidth]{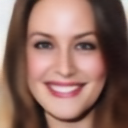}
        \centerline{\scriptsize{EIPNet}}
    \end{minipage}%
    \begin{minipage}[t]{0.09\linewidth}
        \centering
        \includegraphics[width=\textwidth]{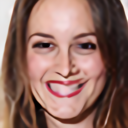}
        \centerline{\scriptsize{KDFSRNet}}
    \end{minipage}%
    \begin{minipage}[t]{0.09\linewidth}
        \centering
        \includegraphics[width=\textwidth]{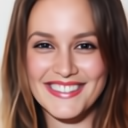}
        \centerline{\scriptsize{UFVNet}}
    \end{minipage}%
    \hspace{2mm}
    {\rotatebox{90}{\scriptsize{~~~~~$56\times56$}}}
     \begin{minipage}[t]{0.09\linewidth}
        \centering
   \includegraphics[width=\textwidth]{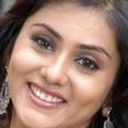}
       \centerline{\scriptsize{HR}}
    \end{minipage}%
    \begin{minipage}[t]{0.09\linewidth}
        \centering
        \includegraphics[width=\textwidth]{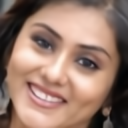}
        \centerline{\scriptsize{SPARNet}}
    \end{minipage}%
    \begin{minipage}[t]{0.09\linewidth}
       \centering
        \includegraphics[width=\textwidth]{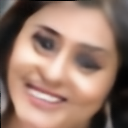}
        \centerline{\scriptsize{EIPNet}}
    \end{minipage}%
    \begin{minipage}[t]{0.09\linewidth}
       \centering
        \includegraphics[width=\textwidth]{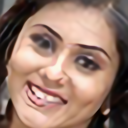}
        \centerline{\scriptsize{KDFSRNet}}
    \end{minipage}%
    \begin{minipage}[t]{0.09\linewidth}
        \centering
        \includegraphics[width=\textwidth]{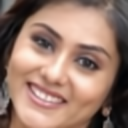}
        \centerline{\scriptsize{UFVNet}}
    \end{minipage}
    \vspace{0.2cm}
%    \caption{A visual comparison on the state-of-the-art methods using original-resolution input and UFVNet. In the left image, models are trained on CelebA-MR8 and in the right image, models are trained on CelebA-MR3. }
    \caption{Visual results by UFVNet and the benchmark OR methods on example images from CelebA-MR8 and Helen-MR8. Results on the left and the right halves are obtained by models trained on CelebA-MR8 and CelebA-MR3, respectively.}

\label{fig:or}
\end{figure*}

%\textbf{Comparison with methods using FR input:}
\textbf{Comparison with FR methods:} 
The state-of-the-art FR methods we compare with include DIC~\cite{ma2020deep}, SISN~\cite{lu2021face}, WIPA~\cite{dastmalchi2022super} and MLGDN~\cite{zhuang2022multi}.
%The state-of-the-art FR methods we compare our proposed UFVNet with are DIC~\cite{ma2020deep}, SISN~\cite{lu2021face}, WIPA~\cite{dastmalchi2022super} and MLGDN~\cite{zhuang2022multi}. 
They all achieve good performance with an 8$\times$ up-scaling factor. Among these works, SISN~\cite{lu2021face} and WIPA~\cite{dastmalchi2022super} train separate models for more than one up-scaling factors and achieve better performance with a 4$\times$ up-scaling factor. As such, we choose the 4$\times$ models for SISN~\cite{lu2021face} and WIPA~\cite{dastmalchi2022super}.
%support more than one upscale factor and achieve better performance with a 4x upscale factor. 
The DIC~\cite{ma2020deep} model provided by the authors is trained on CelebA and we directly use it. The SISN~\cite{lu2021face}, WIPA~\cite{dastmalchi2022super} and MLGDN~\cite{zhuang2022multi} models provided by the authors are not trained on CelebA, and we train them on CelebA by ourselves.
%If there are models for multiple up-scaling factors, we choose the one with the best performance. 
In this experiment, we test the benchmark FR models on both CelebA-MR8 and Helen-MR8. %, and all the output SR images have a resolution of $128 \times 128$. 
We resize each test image to the input resolution 
%(or size) 
of each specific benchmark FR model using bicubic interpolation. For our proposed UFVNet, we train it on CelebA-MR8 and test it on both CelebA-MR8 and Helen-MR8 in this experiment. We resize each test image to the resolution of $128 \times 128$ before inputting it to UFVNet. In addition, we use bicubic interpolation as a baseline FSR approach for comparison.
%For the best performance, we do not modify their networks, and set the upscale factor of each work to the one it achieves the best performance with if there is a choice. 
%We test their fine-tuned models on both CelebA-MR8 and Helen-MR8. The resolution of all the outputing SR images in this experiment is $128 \times 128$. To input multi-resolution test samples into the FR networks, we resize them to the corresponding resolution at the selected upscale factors by using bicubic interpolation. Our proposed UFVNet is trained on CelebA-MR8 in this experiment. We also adopt Bicubic method as the baseline. Different from other methods, the testing set for UFVNet and Bicubic are not resized before inputting.

The quantitative results on CelebA-MR8 and Helen-MR8 are listed in Tab.~\ref{tab:fr}. We clearly observe that UFVNet outperforms all the FR methods and the baseline Bicubic method significantly in both PSNR and SSIM. The results indicate that the FR methods may not super-resolve inputs of varied resolutions satisfactorily by a single model. 
%multi-resolution inputs satisfactorily by a single model. 
We also notice that even the baseline Bicubic method surpasses all the FR methods. This again shows that the FR models are optimized for specific input resolutions but do not generalize well for super-resolving LR images of a large gamut of resolutions.

%It is because that the bicubic method benefits a lot from the higher-resolution inputs (\eg $48\times 48$, $56\times 56$ and $64\times 64$) of high quality already while the FR models are optimized for specific lower input resolutions only. 

%Surprisingly, we notice that the baseline Bicubic method surpasses all the FR methods. It is because that the bicubic method benefits a lot from the higher-resolution inputs (\eg $48\times 48$, $56\times 56$ and $64\times 64$) of high quality already while the FR models are optimized for specific lower input resolutions only. 
%It is because the Bicubic method benefits a lot from the large-resolution inputs, such as $48\times 48$, $56\times 56$ and $64\times 64$, while other FR methods do not benefit from input with a resolution larger than $16\times 16$. 
%We simply remove the testing samples at the highest three resolutions and test the Bicubic method again, its performance drops sharply to the last place, as indicated by the results of Bicubic* in Tab.~\ref{tab:fr}.

We visualize some SR image results produced by different methods in Fig~\ref{fig:fr}. 
%The five rows in Fig~\ref{fig:fr} correspond to %up-scaling factors of $\frac{1}{16}$, $\frac{1}{16}$, $\frac{3}{16}$, $\frac{3}{16}$ 
%LR image sizes of $8 \times 8$, $8 \times8 $, $16 \times 16$, $16 \times 16$ and $56 \times 56$, respectively. 
%It clearly shows the superior performance of UFVNet to the others on the example images. 
It clearly shows higher quality of SR images generated by UFVNet than by the others. Even at a very low input resolution of $8 \times 8$, as shown in the first row, UFVNet still produces identifiable face images. We further show in Fig.~\ref{fig:or_celeba_exp3} that the FR models (\eg DIC, WIPA) generate better results when the input images have the resolutions for which the models are originally trained. For other input resolutions, though, these models do not work optimally. But still, our UFVNet works robustly for the cases shown in Fig.~\ref{fig:or_celeba_exp3}.

\textbf{Comparison with OR methods:}
%The state-of-the-art OR methods we compare our proposed UFVNet with are SPARNet~\cite{chen2020learning}, EIPNet~\cite{Kim2021EdgeAI} and KDFSRNet~\cite{Wang2022PropagatingFP}. 
The state-of-the-art OR methods we compare with include SPARNet~\cite{chen2020learning}, EIPNet~\cite{Kim2021EdgeAI} and KDFSRNet~\cite{Wang2022PropagatingFP}. In this experiment, we make two settings for training and testing all the OR (except KDFSRNet) and UFVNet models: 1) training on CelebA-MR8 and testing on CelebA-MR8 and Helen-MR8, and 2) training on CelebA-MR3 and testing on CelebA-MR8 and Helen-MR8. It should be noted that KDFSRNet is published most recently with trained model but not training code released. Therefore, we use the trained model with a 4$\times$ up-scaling factor for testing.

%We use the multi-scale testing sets to test the OR models trained on the original training sets. Although the LR images are resized to their input resolution, they still fail at most of the inputs. Therefore, we consider re-training the models in the same way as we train UFVNet. In this experiment, we train the OR models and UFVNet on CelebA-MR3 and CelebA-MR8 respectively and test them on CelebA-MR8 and Helen-MR8. It should be noted that KDFSRNet is the latest published paper and has not released the training code, so we use the fine-tuned model with 4x upscale factor provided by the authors for testing.

The quantitative results of this experiment are listed in Tab.~\ref{tab:or} with results for the first settings on the left and the second settings on the right. We clearly observe that UFVNet achieves the best PSNR and SSIM performance among all the models in both settings. The performance gaps between all the OR and the Bicubic methods and UFVNet are obvious. Among all the compared methods, SPARNet achieves performance closer to UFVNet, while EIPNet and KDFSRNet fall far behind. Referring to Tab.~\ref{tab:fr}, we find that EIPNet and KDFSRNet achieve metric values just close to those of the FR methods. This indicates that, without special design, these OR methods still cannot handle inputs of varied resolutions well.
%Referring to Tab.~\ref{tab:fr}, we find that both EIPNet and KDFSRNet achieve metric values close to those of the FR methods. This indicate that, without special design, these OR methods still cannot handle multi-resolution input well. 
%Further, comparing the results in the left and the right parts of Tab.~\ref{tab:or} corresponding to the two experimental settings, we observe a sharper advantage of UFVNet over the other OR methods except KIFSRNet that is not retrained by us. In other words, for input resolutions not seen by the neural networks in the training, UFVNet still performs comparably while the other OR methods may perform obviously worse. This demonstrates the robustness and generalizability of UFVNet.   
Further, we observe from Tab.~\ref{tab:or} that, when the training dataset switches from CelebA-MR8 to CelebA-MR3, SPARNet and EIPNet suffer from larger performance drops than UFVNet. In other words, for input resolutions not seen in the training, UFVNet still performs comparably well while the other OR methods may perform obviously worse. This demonstrates the robustness and extensibility of UFVNet.

We visualize some SR image results produced by the OR methods and UFVNet in Fig.~\ref{fig:or}. %This figure again 
It shows the superior performance of UFVNet in both experimental settings and across a wide range of input resolutions.

\begin{table}[b]
\caption{PSNR and SSIM performance of the %state-of-the-art SASR methods and UFVNet.
benchmark SASR methods and UFVNet. 
In each column, result in \textbf{bold} is the best. }\label{tab:sasr}
\resizebox{.95\columnwidth}{!}{
\begin{tabular}{|c|c|c|c|c|}
\hline
 \multirow{2}{*}{\diagbox{Method}{Dataset}}  & \multicolumn{2}{c|}{CelebA-MR7} & \multicolumn{2}{c|}{Helen-MR7}\\
\cline{2-5}
 & \hspace{-2mm} PSNR$\uparrow$ \hspace{-3mm}
 & \hspace{-1mm} SSIM$\uparrow$ \hspace{-3mm} 
 & \hspace{-1mm} PSNR$\uparrow$ \hspace{-3mm} 
 & \hspace{-1mm} SSIM$\uparrow$ \hspace{-1mm}\\
\hline
\hline
\hspace{-2mm} Bicubic \hspace{-2mm}&28.75 & 0.8231 & 29.03 & 0.8464 \\
\hline
%             MetaSR 26.62  0.8144  26.73, 0.8308
\hspace{-2mm} ArbSR ~\cite{wang2021learning}\hspace{-2mm}&28.77 & 0.8540& 28.82 & 0.8633 \\
\hline
\hspace{-2mm} LIIF ~\cite{chen2021learning}\hspace{-2mm}& 29.90 & 0.8449 & 30.13 & 0.8553 \\
\hline
\hspace{-2mm} UFVNet(Ours) \hspace{-2mm} & \textbf{30.43} & \textbf{0.8606}&\textbf{30.48} & \textbf{0.8663} \\
\hline
\end{tabular}}
\end{table}

\begin{table*}[htb] \centering
\caption{Quantitative comparison of the six models on the CelebA and the Helen datasets.
In each column, result in \textbf{bold} is the best.}
\vspace{0.2cm}
\label{tab:Ablation1}
\begin{tabular}{|c|c|c|c|c|c|c|c|c|}
\hline
 \multirow{3}{*}{\diagbox{Method}{Dataset}}
 &\multicolumn{4}{c|}{Trained on CelebA-MR8}&\multicolumn{4}{c|}{Trained on CelebA-MR3}\\
 \cline{2-9}
 & \multicolumn{2}{c|}{CelebA-MR8}& \multicolumn{2}{c|}{Helen-MR8}  &\multicolumn{2}{c|}{CelebA-MR8} & \multicolumn{2}{c|}{Helen-MR8}\\
\cline{2-9}
 & \hspace{-2mm} PSNR$\uparrow$ \hspace{-3mm}
 & \hspace{-3mm} SSIM$\uparrow$ \hspace{-3mm} 
 & \hspace{-3mm} PSNR$\uparrow$ \hspace{-3mm} 
 & \hspace{-1mm} SSIM$\uparrow$ \hspace{-1mm}
 & \hspace{-2mm} PSNR$\uparrow$ \hspace{-3mm}
 & \hspace{-3mm} SSIM$\uparrow$ \hspace{-3mm} 
 & \hspace{-3mm} PSNR$\uparrow$ \hspace{-3mm} 
 & \hspace{-1mm} SSIM$\uparrow$ \hspace{-1mm}\\
\hline
\hline
\hspace{-3mm} UFVNet-NW \hspace{-2mm} &29.25 & 0.8299 & 29.19 & 0.8410 & 28.89 & 0.8161 & 29.13 & 0.8378\\
\hline
\hspace{-2mm} UFVNet-FM \hspace{-2mm}& 29.35&0.8310 & 29.45 & 0.8320 & 24.27& 0.7612& 24.21& 0.7820 \\
\hline
\hspace{-2mm} OneSRG\hspace{-2mm}& 29.20&0.8201 & 29.34 & 0.8406 & 28.73& 0.8112& 28.90& 0.8325\\
\hline
\hspace{-2mm} TwoSRG\hspace{-2mm}&29.40 & 0.8283 & 29.46 & 0.8430 &28.80 & 0.8137 & 29.04 & 0.8342 \\
\hline
\hspace{-2mm} FourSRG\hspace{-2mm}& \textbf{29.48} & 0.8317 &  \textbf{29.51} & 0.8436  &28.64 & 0.8136 & 28.71 & 0.8346\\
\hline
\hspace{-2mm} UFVNet(Ours)  &29.46 &  \textbf{0.8321}&29.49 & \textbf{0.8446} &\textbf{29.16}&\textbf{0.8241} &\textbf{29.46}&\textbf{0.8453}\\
\hline
\end{tabular}
\end{table*}

\begin{figure}[htbp]
\centering
    {\rotatebox{90}{\scriptsize{~~~~~$32\times32$}}}
    \begin{minipage}[t]{0.19\linewidth}
        \centering
        \includegraphics[width=\textwidth]{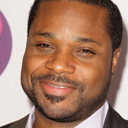}
    \end{minipage}%
    \begin{minipage}[t]{0.19\linewidth}
        \centering
        \includegraphics[width=\textwidth]{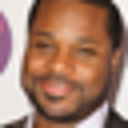}
    \end{minipage}%
    \begin{minipage}[t]{0.19\linewidth}
        \centering
        \includegraphics[width=\textwidth]{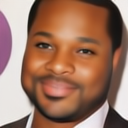}
    \end{minipage}%
    \begin{minipage}[t]{0.19\linewidth}
        \centering
        \includegraphics[width=\textwidth]{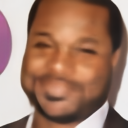}
    \end{minipage}%
    \begin{minipage}[t]{0.19\linewidth}
        \centering
        \includegraphics[width=\textwidth]{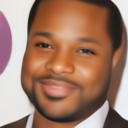}
    \end{minipage}
    {\rotatebox{90}{\scriptsize{~~~~~$24\times24$}}}
    \begin{minipage}[t]{0.19\linewidth}
        \centering
        \includegraphics[width=\textwidth]{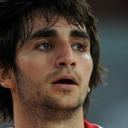}
    \end{minipage}%
    \begin{minipage}[t]{0.19\linewidth}
        \centering
        \includegraphics[width=\textwidth]{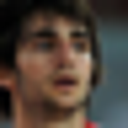}
    \end{minipage}%
    \begin{minipage}[t]{0.19\linewidth}
        \centering
        \includegraphics[width=\textwidth]{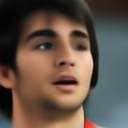}
    \end{minipage}%
    \begin{minipage}[t]{0.19\linewidth}
        \centering
        \includegraphics[width=\textwidth]{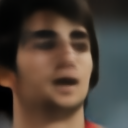}
    \end{minipage}%
    \begin{minipage}[t]{0.19\linewidth}
        \centering
        \includegraphics[width=\textwidth]{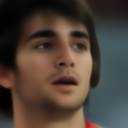}
    \end{minipage}
    {\rotatebox{90}{\scriptsize{~~~~~$16\times16$}}}
    \begin{minipage}[t]{0.19\linewidth}
        \centering
        \includegraphics[width=\textwidth]{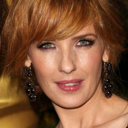}
        \centerline{\scriptsize{HR}}
    \end{minipage}%
    \begin{minipage}[t]{0.19\linewidth}
        \centering
        \includegraphics[width=\textwidth]{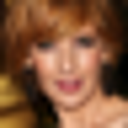}
        \centerline{\scriptsize{Bicubic}}
    \end{minipage}%
    \begin{minipage}[t]{0.19\linewidth}
        \centering
        \includegraphics[width=\textwidth]{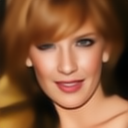}
        \centerline{\scriptsize{ArbSR}}
    \end{minipage}%
    \begin{minipage}[t]{0.19\linewidth}
        \centering
        \includegraphics[width=\textwidth]{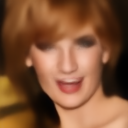}
        \centerline{\scriptsize{LIIF}}
    \end{minipage}%
    \begin{minipage}[t]{0.19\linewidth}
        \centering
        \includegraphics[width=\textwidth]{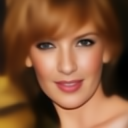}
        \centerline{\scriptsize{UFVNet}}
    \end{minipage}
    \hspace{0.1cm}
    \vspace{0.1cm}
%    \caption{A visual comparison on the state-of-the-art SASR methods and UFVNet. }
    \caption{Visual results by UFVNet and the benchmark SASR methods on example images from CelebA-MR7 and Helen-MR7.}
    \label{fig:sasr}
\end{figure}

%\textbf{Comparison with methods using SASR input:}
\textbf{Comparison with SASR methods:}
We have found no FSR algorithm that processes inputs of varied resolutions with a single model. But still, we compare UFVNet with two representative SASR models, \ie ArbSR~\cite{wang2021learning} and LIIF~\cite{chen2021learning}, that were recently proposed for generic image super-resolution. For fairness of comparison, we train ArbSR, LIIF and UFVNet all on CelebA-MR7 and test them all on CelebA-MR7 and Helen-MR7. Note that we exclude the very low resolution, \ie $\frac{1}{16}$ the original HR, from our training and testing datasets. This is to avoid biased comparison as ArbSR and LIIF produce $16 \times$ SR images of very poor quality on our datasets. For the training of ArbSR, LR images of various sizes are used as input and their corresponding $128 \times 128$ HR images are used as ground-truth. For the training of LIIF, ground-truth references are re-generated from training images by random scaling, which are down-sized to $16 \times 16$ images for input.

The quantitative results are listed in Tab.~\ref{tab:sasr}. We clearly observe that UFVNet outperforms the SASR methods and the baseline Bicubic method significantly in both PSNR and SSIM. In addition, we visualize some SR image results of different methods in Fig.~\ref{fig:sasr}. From these examples, we observe that our proposed algorithm achieves clearly better quality of SR images than the others. This reflects the fact that the SASR methods are optimized for generic image SR but not so for specific FSR tasks.

\subsection{Ablation Study}

The success of our unified framework in processing varied input face image resolutions is mainly due to two components of design, \emph{i.e.} multiple anchor resolutions feature extraction and adaptive weight regression. In order to verify the effectiveness of these two key components, we conduct an ablation study in this subsection. 

%In order to validate these components, we make five additional networks.
We make five additional models for the ablation study.
UFVNet-NW is our network with the weight regression branch removed and the weights of the three encoder-decoder branches all fixed to 1. UFVNet-FM calculates the weights of the three encoder-decoder branches directly using the formulae in Eq.\ref{eqn:weight}. OneSRG, TwoSRG and FourSRG are the networks using one, two and four encoder-decoder branches, respectively. Besides, our proposed UFVNet, namely ThreeSRG, consists of three encoder-decoder branches. 

We train all these models on the two training sets and test them on CelebA-MR8 and Helen-MR8. The results are shown in Tab.~\ref{tab:Ablation1}. We observe: 1) UFVNet significantly outperforms UFVNet-NW, UFVNet-FM and OneSRG, showing the effectiveness of both adaptive weight regression and multiple anchor resolutions feature extraction; 2) UFVNet (ThreeSRG) outperforms OneSRG, TwoSRG and FourSRG in overall performance, indicating that three encoder-decoder branches is the best choice.

\section{Conclusion}
In this work, we have proposed UFVNet, a unified framework to super-resolve face images of varied low resolutions. To the best of our knowledge, this is the first published framework of this type for the specific problem of FSR. 
%To the best of our knowledge, we make the first attempt to address this problem \textcolor{red}{in FSR field}. 
The other existing FSR algorithms usually train a specific model for a specific low input resolution for optimal results but do not generalize well for super-resolving face images of varied resolutions. 
%Though some of them may be trained as general models, but the original designs were not oriented for varied input resolutions and the results are sub-optimal. 

The proposed UFVNet consists of three anchor branches each meant to work optimally for a pre-defined input resolution and a weight regression branch giving weights to combine the encoded features of the three anchor branches. Finally, the features are fused and fed to a final decoder to generate the final SR result. As experimentally demonstrated, the proposed FSR algorithm achieves superior performance over a large gamut of input resolutions by a single framework.

\bibliographystyle{IEEEtran}
\bibliography{UFVNet}

\end{document}